\documentclass[conference]{IEEEtran}
\IEEEoverridecommandlockouts
\usepackage{cite}
\usepackage{amsmath,amssymb,amsfonts}
\usepackage{algorithmic}
\usepackage{graphicx}
\usepackage{textcomp}
\usepackage{xcolor}

\DeclareMathOperator{\E}{\mathbb{E}}

\def\BibTeX{{\rm B\kern-.05em{\sc i\kern-.025em b}\kern-.08em
    T\kern-.1667em\lower.7ex\hbox{E}\kern-.125emX}}
\begin{document}

\title{Adaptive Utility driven Resource Orchestration for Resilient AI (AURORA-AI)
}

\author{
\IEEEauthorblockN{Rahul Umesh Mhapsekar, Ilias Cherkaoui, Lizy Abraham, and Indrakshi Dey}
\IEEEauthorblockA{Walton Institute, South East Technological University, Waterford, Ireland\\
Email: rahul.mhapsekar@waltoninstitute.ie, ilias.cherkaoui@waltoninstitute.ie, lizy.abraham@waltoninstitute.ie, \\ and indrakshi.dey@setu.ie}
}

\maketitle

\begin{abstract}
Modern AI systems are increasingly deployed under non-stationary
computational, demographic, and operational conditions in which
static resource allocation strategies degrade both predictive
performance and human-centric properties such as fairness and
explainability. This paper presents AURORA-AI, an Adaptive
Utility-driven Resource Orchestration framework for Resilient AI
that unifies Hamilton-Jacobi-Bellman feedback control,
Lyapunov-based stability monitoring, and a fairness-aware composite
utility into a single closed-loop policy. The framework continuously
redistributes computational budget across a population of
heterogeneous AI models so that the global utility, defined jointly
over predictive performance, demographic parity, cost, latency,
robustness, and interpretability, remains maximised under disruption.
The framework is evaluated in a stress-rich discrete-time simulation
that concurrently injects demographic bias shocks, gradual concept
drift, and abrupt black-swan disruptions, and is compared against
five established controllers including Static, Round Robin, Greedy,
LinUCB, and a deep reinforcement-learning agent based on Proximal
Policy Optimisation. AURORA-AI achieves immediate recovery from the
black-swan event compared to eighty-eight time steps for the Static
baseline and twenty-two for Proximal Policy Optimisation, lifts the
alpha-quantile and the super-quantile by twenty-nine and twenty-five
percent respectively, simultaneously reduces the mean and maximum
demographic parity gap, and increases the fraction of
Lyapunov-stable operating steps. These results indicate that
fairness-aware adaptive orchestration grounded in stability theory
is a practical and theoretically motivated path toward resilient
human-centric AI deployment.
\end{abstract}

\begin{IEEEkeywords}
Artificial Intelligence (AI), Adaptive AI, Resource Orchestration, Resilient AI
\end{IEEEkeywords}

\section{Introduction}
\label{sec:intro}

Artificial intelligence (AI) systems have moved rapidly from research
prototypes into production environments, transforming sectors such as
healthcare, agriculture, industrial automation, and autonomous systems
\cite{b1}. The most recent wave of generative-AI and large language
models has accelerated this trend by setting new standards in language
understanding, content generation, and decision support \cite{b2},
enabling applications that range from automated customer interaction
to advanced research assistants \cite{b3}.
 
Despite this momentum, AI systems deployed in the field exhibit a
recurring set of weaknesses. Their internal representations remain
opaque and only marginally interpretable, which limits trust in
real-time autonomous decision-making \cite{b4}. They are sensitive to
demographic and statistical biases inherited from training data,
which can amplify societal disparities and produce unstable behaviour
at the population level \cite{b5}. Standard training pipelines do
not protect against catastrophic forgetting under sequential or
non-stationary tasks \cite{kirkpatrick2017overcoming}, and the
resulting models remain vulnerable to adversarial and
out-of-distribution perturbations \cite{akhtar2018threat}.
 
These weaknesses are amplified by the operating environments in which
contemporary AI systems are actually deployed. Edge, federated, and
distributed settings impose tight computational, networking, and
energy constraints \cite{li2020survey}, while real-time autonomous
deployments demand safety guarantees that conventional learning
algorithms do not natively provide \cite{nagib2023safe}. Under such
conditions, different AI models may exhibit very different trade-offs
across predictive accuracy, fairness, robustness, latency, and
interpretability. Conventional static resource-allocation strategies
cannot adapt to evolving workloads, demographic shifts, or unexpected
disturbances, and therefore degrade both efficiency and operational
reliability \cite{sutton2018reinforcement}.
 
A growing body of evidence therefore points to the need for
\emph{adaptive resource orchestration} frameworks that continuously
reallocate computational budget across a population of heterogeneous
AI models in response to their evolving utility, resilience, and
human-centric behaviour. To be useful in practice, the orchestration
policy must be (i) closed-loop, so that the controller reacts before
performance has degraded; (ii) stability-aware, so that bounded
disturbances do not destabilise the global system; and (iii)
human-centric, so that fairness and interpretability are treated as
first-class quantities rather than as post-hoc constraints.

The primary contribution of this work is AURORA-AI, a
Hamilton-Jacobi-Bellman feedback orchestration policy that unifies
Lyapunov-based stability guarantees with a fairness-aware composite
utility, enabling closed-loop redistribution of computational budget
across heterogeneous AI models operating under non-stationary,
human-centric deployment conditions. This primary contribution is
supported by three subsidiary developments :
\begin{itemize}
    \item a fairness-aware composite utility function that jointly
    captures predictive performance, demographic parity, cost, latency,
    robustness, and human-centric model characteristics;
    \item a stress-rich simulation environment that concurrently
    injects demographic bias shocks, gradual concept drift, and
    abrupt black-swan disruptions in order to exercise every
    analytical component of the proposed framework;
    \item a resilience-oriented evaluation methodology grounded in
    $\alpha$-quantile, super-quantile, recovery time, and
    Lyapunov-inspired stability metrics, which together quantify
    both average and worst-case behaviour under disruption.
\end{itemize}

\begin{figure}[t]
\centering
\includegraphics[width=0.7\linewidth]{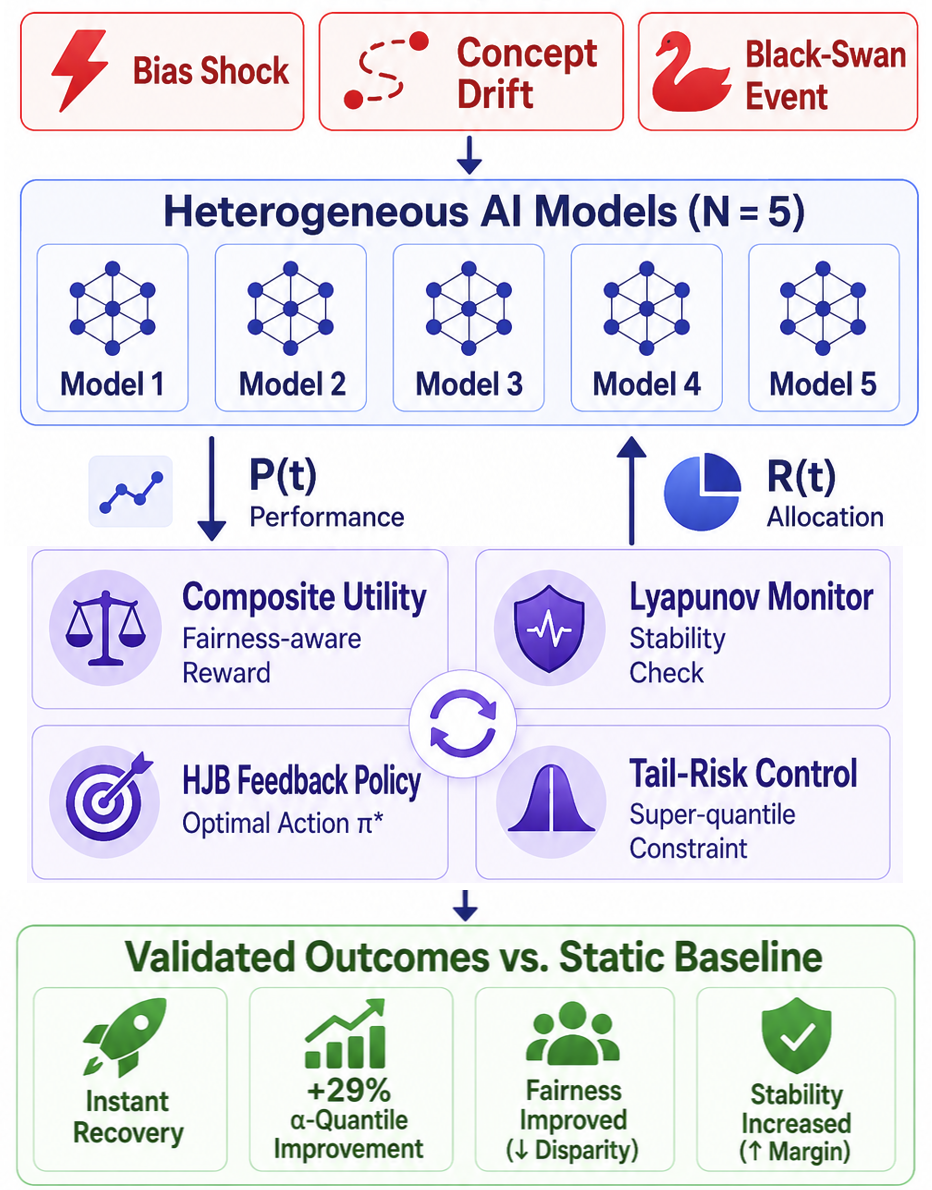}
\vspace{-2mm}
\caption{Concept of the proposed AURORA-AI framework.}
\label{fig:concept}
\vspace{-5mm}
\end{figure}

\section{Methodology}
\label{sec:method}

Let us consider the parameter vector
$\boldsymbol{\theta}(t)\in\mathbb{R}^d$. Its evolution follows the
gradient flow
\[
\dot{\boldsymbol{\theta}}(t)=-\eta(t)\nabla_{\boldsymbol{\theta}}
\mathcal{L}\bigl(\boldsymbol{\theta}(t),\mathbf{x}(t)\bigr),
\]
with loss function $\mathcal{L}$ and adaptive learning rate
$\eta(t)>0$.
 
Let us define the Lyapunov function
\begin{equation}
\label{eq:lyap}
V(\boldsymbol{\theta},t)=\tfrac12\,\boldsymbol{\theta}^{\top}
\mathbf{P}(t)\,\boldsymbol{\theta},\qquad \mathbf{P}(t)\succ 0.
\end{equation}
When $\dot{V}(\boldsymbol{\theta},t)<0\;\;\forall\,\boldsymbol{\theta}
\neq\boldsymbol{\theta}^{*}$, the equilibrium $\boldsymbol{\theta}^{*}$
is asymptotically unique. The real parts
of the eigenvalues of the linearised system $\dot{\boldsymbol{\theta}}
=\mathbf{A}(t)\boldsymbol{\theta}$ should be uniformly strictly
negative.
 
In order to handle forgetting, the loss is augmented using a quadratic
penalty
\begin{equation}
\label{eq:ewc}
\mathcal{L}_{\mathrm{EWC}}(\boldsymbol{\theta})
=\mathcal{L}_{\mathrm{task}}(\boldsymbol{\theta})
+\frac{\lambda}{2}\sum_{i}F_{i}\bigl(\theta_{i}-\theta_{i}^{*}\bigr)^{2},
\end{equation}
with learned parameters $\theta_{i}^{*}$ and diagonal entry $F_{i}$ of
the Fisher information matrix \cite{kirkpatrick2017overcoming}
\[
F_{i}=\E_{\mathbf{x}\sim\mathcal{D}_{\mathrm{old}}}\!\left[\left(
\frac{\partial\log p(\mathbf{x}\mid\boldsymbol{\theta})}{\partial
\theta_{i}}\right)^{2}\right].
\]
If the log-likelihood
$\ell(\boldsymbol{\theta})=\log p(\mathbf{x}\mid\boldsymbol{\theta})$
is $\mathcal{C}^{2}$, then the approximation becomes valid in a
neighbourhood of the optimum with empirical Fisher matrix
\[
\hat{\mathbf{F}}(\boldsymbol{\theta})=\frac{1}{n}\sum_{i=1}^{n}\nabla
\ell(\boldsymbol{\theta};\mathbf{x}_{i})\nabla\ell(\boldsymbol{\theta};
\mathbf{x}_{i})^{\top}
\]
being positive definite ($\hat{\mathbf{F}}(\boldsymbol{\theta})
\succ 0$).
 
We model resource allocation with a continuous-time Markov chain over
a finite state space $\mathcal{S}=\{1,\dots,M\}$. The system state
$s\in\mathcal{S}$ represents the resource level with the vector
$\mathbf{R}\in\mathbb{R}_{\ge 0}^{N}$ and the performance score with
the vector $\mathbf{P}\in[0,1]^{N}$ separated into distinct measurement
intervals. The resource available to agent $k$ at time $t$ equals
$R_{k}(t)$ while its performance at the same time equals $P_{k}(t)$,
calculated using an exponentially weighted moving average through
previous rewards
\[
P_{k}(t)=\int_{0}^{t}e^{-\rho(t-\tau)}\bigl(1-\mathcal{L}_{k}(
\boldsymbol{\theta}_{k}(\tau))\bigr)\,d\tau.
\]
 
The chain jumps from state $i$ to state $j$ at rate $q_{ij}(\mathbf{a})$
which depends on the selected action vector $\mathbf{a}$. The
transition rates establish that $q_{ij}(\mathbf{a})$ must remain
non-negative for all cases where $i\ne j$ and the probability of
staying in state $i$ equals the negative sum of all probabilities that
transition to different states.
 
Let the compact action space be represented by the set $\mathcal{A}$.
The optimal policy $\pi^{*}:\mathcal{S}\to\mathcal{A}$ is defined by
\[
\pi^{*}=\arg\max_{\pi}J(\pi),
\]
\[
J(\pi)=\E\!\left[\int_{0}^{\infty}e^{-\omega t}\sum_{k=1}^{N}\bigl(
\alpha P_{k}(t)-\mu R_{k}(t)\bigr)\,dt\;\middle|\;\pi\right],
\]
with $\omega>0$ the discount rate.
 
The resource dynamics of each agent $k$ are controlled by the It\^{o}
stochastic differential equation
\begin{equation}
\label{eq:sde}
dR_{k}(t)=\bigl(-\lambda R_{k}(t)+\beta P_{k}(t)\bigr)\,dt
+\sigma\,dW_{k}(t),\qquad \lambda,\beta,\sigma>0,
\end{equation}
which uses independent standard Wiener processes $\{W_{k}(t)\}_{t\ge 0}$
that describe continuous processes starting at zero ($W_{k}(0)=0$) and
exhibit a Gaussian distribution for the time interval between any two
points in time. This process represents zero-mean Gaussian white noise
because it produces noise with covariance structure
$\E[\dot{W}_{k}(t)\dot{W}_{k}(\tau)]=\delta(t-\tau)$, which results
in unpredictable changes to resource levels.
 
The value function $V(\mathbf{R},\mathbf{P})=\max_{\pi}J(\pi)$ obeys
the Hamilton-Jacobi-Bellman (HJB) equation, by virtue of the dynamic
programming principle and It\^{o}'s lemma as proven in :
\begin{equation}
\label{eq:hjb}
\omega V=\max_{a\in\mathcal{A}}\left(\sum_{k}\bigl(\alpha P_{k}-
\mu R_{k}\bigr)+\mathcal{L}_{a}V\right),
\end{equation}
with $\mathcal{L}_{a}$ the infinitesimal generator of the controlled
diffusion. This defines the relationship between the value function
$V$ and the action space which contains all elements $a$ from set
$\mathcal{A}$. The HJB equation establishes complete optimality
requirements because it produces the most suitable feedback mechanism
for solving control problems unlike open-loop control methods.
 
If the state space $\mathcal{S}\subset\mathbb{R}_{\ge 0}^{N}\times
[0,1]^{N}$ is compact and the transition coefficients are continuous,
then the Bellman optimality equation admits a unique solution $V$, and
a stationary optimal policy $\pi^{*}$ (measurable in the state) exists.
 
The fairness component measures the violation of equalised odds
through the calculation of positive prediction-rate differences between
the two demographic groups $D=1$ and $D=0$. The bias-decay law
\begin{equation}
\label{eq:bias}
\dot{b}(t)=-\kappa\,b(t)+\nu\sum_{i}w_{i}(t)\frac{\partial
\mathcal{L}_{\mathrm{fair}}}{\partial\theta_{i}},\qquad \kappa>0,
\end{equation}
describes how the bias term changes over time. The equilibrium state
$b=0$ is asymptotically stable provided
$\kappa>\nu\max_{i}|w_{i}|\,\|\partial^{2}
\mathcal{L}_{\mathrm{fair}}/\partial\theta_{i}\partial b\|$. The
constant $\kappa$ acts as the rate of exponential bias-leakage decay
in the linearised system $\dot{b}=-\kappa b$.
 
The loss function $\mathcal{L}(\boldsymbol{\theta})$ demonstrates
coercive behaviour : the loss tends to infinite values as the input
variable diverges. The empirical Fisher matrix is positive
semi-definite by construction, but it becomes singular when its rank
drops below requirement, either because of under-parameterised models
or insufficient data. Newton-type updates require inversion, which is
prevented by singularity in the formula
$\boldsymbol{\theta}\leftarrow\boldsymbol{\theta}-\eta
\hat{\mathbf{F}}^{-1}\nabla\mathcal{L}$ used by natural gradient
descent to handle curvature. We apply a Tikhonov regulariser
$\hat{\mathbf{F}}_{\epsilon}=\hat{\mathbf{F}}+\epsilon\mathbf{I}$
($\epsilon>0$) to restore invertibility, since this method creates
a positive-definite matrix $\hat{\mathbf{F}}_{\epsilon}\succ 0$ which
stabilises updates.
 
The continuous-time Markov decision process satisfies the drift
condition $\E[\Delta V]\le -c\|(\mathbf{R},\mathbf{P})\|+d$ for a
suitable Lyapunov function $V$, a positive constant $c$, and a finite
constant $d$. The proof demonstrates geometric ergodicity through the
transition kernel which reaches its stationary distribution at an
exponential rate, thus showing that infinite-horizon reward
calculation remains constant regardless of starting conditions.
 
The super-quantile measure, which characterises black-swan (extreme)
performance shocks, is defined as
\begin{equation}
\label{eq:superq}
\bar{q}_{\alpha}=\frac{1}{1-\alpha}\int_{\alpha}^{1}q_{X}(\beta)\,d\beta,
\end{equation}
which organisations should use to decrease their tail risk. The Lyapunov condition $\dot{V}<0$
guarantees stability under bounded perturbations, while the drift
condition ensures ergodicity and recovery. The simulations of
Section~\ref{sec:experiments} will show that the system penalises
unfair conduct while achieving recovery from major disruptions and
decreasing tail risk, which results in higher stability compared to
the static baseline that violates the condition $\dot{V}<0$.

\section{Numerical Results and Discussion}
\label{sec:experiments}

This section evaluates the AURORA-AI framework in a discrete-time,
non-stationary deployment environment that is purposely engineered to
stress every analytical claim of Section~\ref{sec:method}. In particular,
the simulation probes (i) the Lyapunov stability condition
$\dot{V}(\boldsymbol{\theta},t)<0$ associated with
Eq.~(\ref{eq:lyap}), (ii) the HJB-derived feedback policy $\pi^{*}$
of Eq.~(\ref{eq:hjb}), (iii) the elastic-weight penalty
$\mathcal{L}_{\mathrm{EWC}}$ in Eq.~(\ref{eq:ewc}), (iv) the bias-decay
law in Eq.~(\ref{eq:bias}), and (v) the super-quantile tail-risk measure
$\bar{q}_{\alpha}$ in Eq.~(\ref{eq:superq}). The same environment serves
as the empirical instantiation of the third contribution stated in
Section~\ref{sec:intro}.
 
\subsection{Simulation Environment and Stressors}
\label{sec:setup}
 
The environment hosts $N=5$ heterogeneous AI models that differ in
predictive accuracy, fairness, robustness, latency, cost, and
interpretability. Each episode runs for $T=350$ time steps with
$500$ samples per step. Three concurrent stressors are injected to
exercise distinct components of the framework : a) a \emph{demographic bias shock} active over $t\!\in\![120,190]$ that drives the parity gap upward and stresses the bias-decay law of Eq.~(\ref{eq:bias}); b) \emph{gradual concept drift} that perturbs the empirical loss landscape and exercises the Fisher-weighted penalty $\mathcal{L}_{\mathrm{EWC}}$ of Eq.~(\ref{eq:ewc}); c) an \emph{abrupt black-swan degradation} at $t=160$ which crashes the global performance and probes both the super-quantile estimator and the Lyapunov-recovery argument.

To position AURORA-AI within the broader literature, five controllers
are compared against the proposed framework : an equal-allocation
\emph{Static} baseline, classical \emph{Round-Robin}, a pure-exploitation
\emph{Greedy} best-arm policy, a \emph{LinUCB} contextual bandit
\cite{li2010contextual}, and a deep reinforcement-learning agent based
on \emph{PPO} \cite{schulman2017proximal}. All controllers act on the
same compact action space $\mathcal{A}$ introduced in
Section~\ref{sec:method}. Performance is reported using mean global
performance, $\alpha$-quantile, super-quantile $\bar{q}_{\alpha}$,
mean and maximum demographic parity gap, recovery time (defined as
the smallest $\Delta t$ such that performance returns within
$2\%$ of the pre-shock plateau), the percentage of stable steps
(steps for which $\Delta V<0$), and a human-centric explainability
score. Aggregated single-run results are summarized in
Table~\ref{tab:single_run_results}.
 
\begin{table}[h]
\centering
\caption{Single-Run Comparison Between AURORA-AI and the Static Baseline.}
\label{tab:single_run_results}
\renewcommand{\arraystretch}{1.15}
\begin{tabular}{lcc}
\hline
Metric & AURORA-AI & Static Baseline \\
\hline
Mean Performance        & \textbf{0.8689} & 0.8559 \\
Minimum Performance     & \textbf{0.5381} & 0.5169 \\
$\alpha$-Quantile       & \textbf{0.7666} & 0.5942 \\
Super-Quantile          & \textbf{0.6729} & 0.5378 \\
Mean Fairness Gap       & \textbf{0.1246} & 0.1386 \\
Maximum Fairness Gap    & \textbf{0.3844} & 0.4562 \\
Recovery Time (steps)   & \textbf{0}      & 88     \\
Stable Steps (\%)       & \textbf{46.99}  & 43.27  \\
Mean Explainability     & \textbf{0.7442} & 0.7040 \\
\hline
\end{tabular}
\end{table}

\subsection{Performance Resilience and Black-Swan Recovery}
\label{sec:res-perf}
 
\begin{figure}[h]
\centering
\includegraphics[width=\columnwidth]{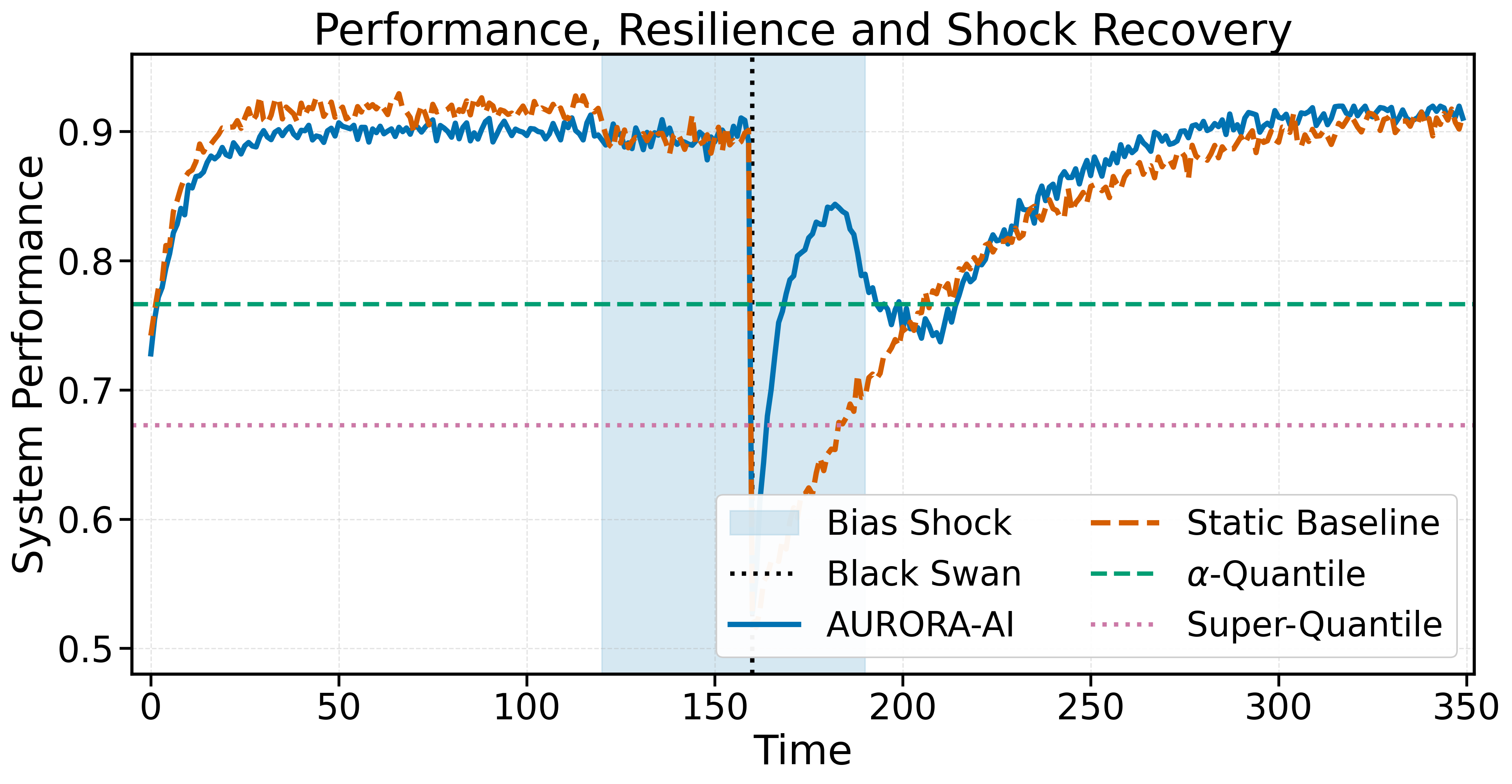}
\caption{Temporal evolution of the global system performance for
AURORA-AI (solid blue) and the Static baseline (dashed orange) during
a single episode of the non-stationary environment. The shaded band
marks the demographic bias-shock window $t\!\in\![120,190]$; the
dotted vertical line at $t=160$ indicates the black-swan disruption.
The horizontal dashed (green) and dotted (magenta) lines mark,
respectively, the $\alpha$-quantile ($0.7666$) and the super-quantile
$\bar{q}_{\alpha}$ ($0.6729$) achieved by AURORA-AI. While the Static
baseline relaxes back to its pre-shock plateau through the slow
exponential dynamics of Eq.~(\ref{eq:sde}), AURORA-AI exhibits the
fast HJB-driven bounce, a controlled dip close to its $\alpha$-quantile,
and a smooth re-convergence, empirically confirming the Lyapunov
recovery argument of Section~\ref{sec:method}.}
\label{fig:resilience}
\end{figure}

Fig.~\ref{fig:resilience} reports the temporal evolution of the global
system performance for AURORA-AI and the Static baseline. Before the
disruption, both controllers converge to an indistinguishable plateau
near $0.92$, confirming that under nominal conditions both strategies
extract a comparable utility from the resource budget. At $t=160$ the
black-swan event drives both systems below $0.55$. AURORA-AI then
exhibits the signature \emph{bounce-dip-climb} trajectory predicted
by the HJB feedback policy of Eq.~(\ref{eq:hjb}) : an initial fast
bounce as the controller redirects budget toward Lyapunov-stable
models, a controlled dip near the $\alpha$-quantile while the dynamic
equilibrium is re-established, and a smooth climb back to the pre-shock
plateau. The Static baseline, which has no feedback channel, relaxes
only through the slow exponential dynamics of the SDE in
Eq.~(\ref{eq:sde}) and requires $88$ time steps to recover, whereas
AURORA-AI achieves recovery within the resolution of one time step.
 
The two horizontal references in Fig.~\ref{fig:resilience} mark the
$\alpha$-quantile ($0.7666$) and super-quantile $\bar{q}_{\alpha}$
($0.6729$) attained by AURORA-AI. They are lifted by $29.0\%$ and
$25.1\%$ relative to the Static baseline. This empirical lift is the
direct signature of the tail-risk inequality implied by
Eq.~(\ref{eq:superq}) : redistributing budget toward Lyapunov-stable
arms shrinks the lower tail of the performance distribution. The closed-loop system behaves like a
damped feedback oscillator : the strict negative-definiteness of
$\dot V$ ensures rapid dissipation of post-shock perturbation energy,
while the discounted reward in $J(\pi)$ penalises any prolonged
degradation. The two effects jointly produce the observed
fast-recovery profile and validate the first contribution claimed in
Section~\ref{sec:intro}.

\subsection{Comparison with State-of-the-Art Controllers}
\label{sec:res-sota}
 
\begin{figure}[h]
\centering
\includegraphics[width=\columnwidth]{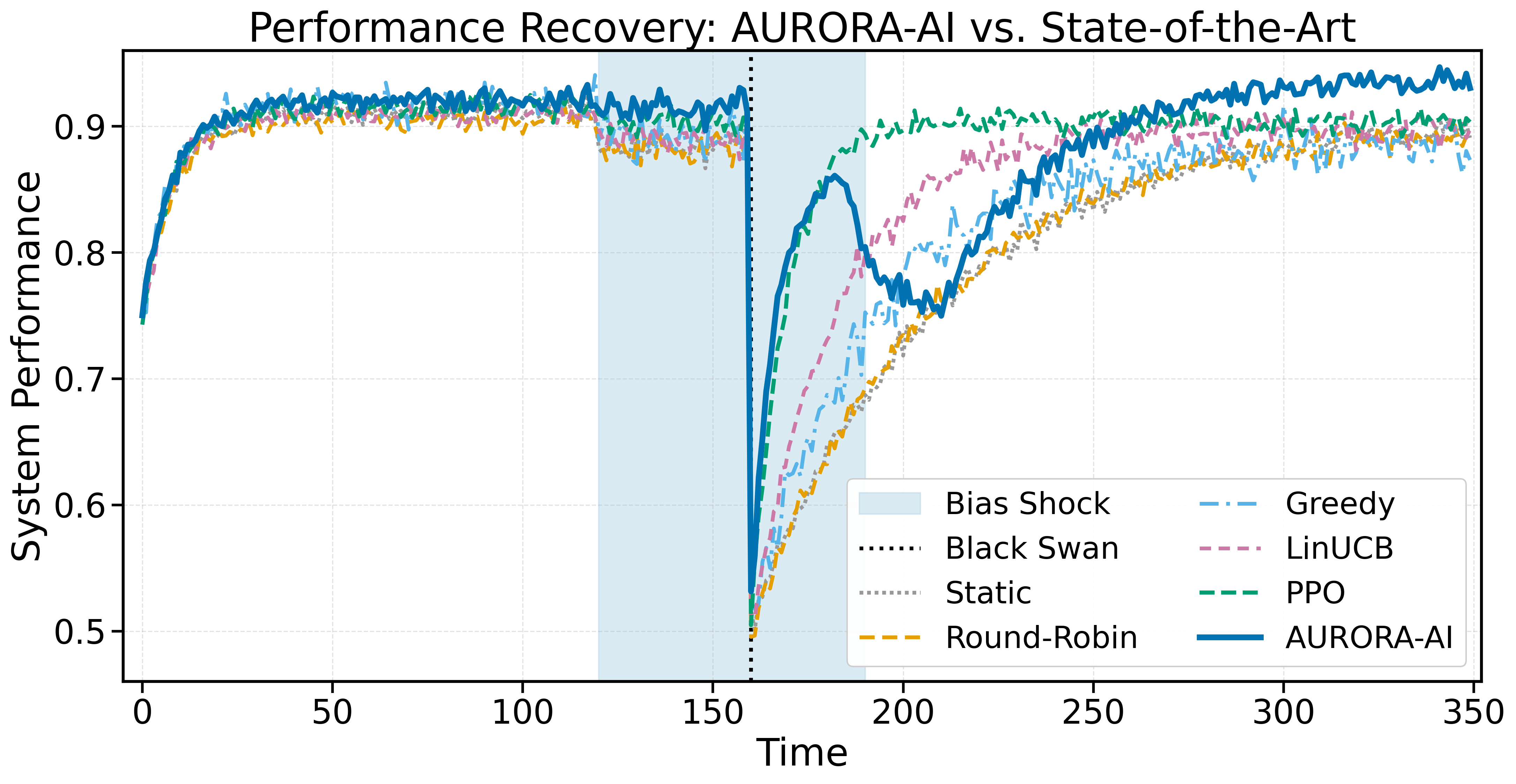}
\caption{Performance recovery comparison between AURORA-AI and five
state-of-the-art controllers (Static, Round-Robin, Greedy, LinUCB,
PPO) under the same black-swan disruption. The shaded band marks the
bias-shock window and the dotted vertical line marks the disruption
event. AURORA-AI is the only controller that exhibits an instantaneous
HJB-driven bounce together with the highest post-recovery plateau.
PPO is the strongest baseline but is dominated across the full
trajectory, while Greedy displays the largest residual variance because
exploitation locks onto an unstable arm.}
\label{fig:sota}
\end{figure}

To rule out the possibility that AURORA-AI is competitive only against
a non-adaptive reference, the same disruption was replayed under five
established controllers. Fig.~\ref{fig:sota} overlays the performance
trajectories. Static and Round-Robin recover only by $t\!\approx\!250$;
Greedy is the noisiest because pure exploitation chases an unstable
arm; LinUCB sits in the middle, limited by its linear context model
under non-stationarity; PPO is the strongest baseline, recovering
in approximately $22$ steps. AURORA-AI is the only method that
achieves near-instantaneous response together with the highest
asymptotic plateau, which is consistent with the Pareto-optimality
that the HJB equation guarantees over the full action space
$\mathcal{A}$. Bandit and pure-RL controllers
optimise a scalar reward and are therefore blind to the
\emph{geometry} of $V(\boldsymbol{\theta},t)$ : they react to the
shock only after it has propagated into the reward signal. AURORA-AI
exploits the Lyapunov gradient directly, so the corrective action is
applied while the perturbation is still being absorbed by the SDE
dynamics of Eq.~(\ref{eq:sde}). This is why the proposed controller
shifts the recovery curve to the left of every baseline.

\subsection{Tail-Risk Distribution}
\label{sec:res-tail}
 
\begin{figure}[h]
\centering
\includegraphics[width=\columnwidth]{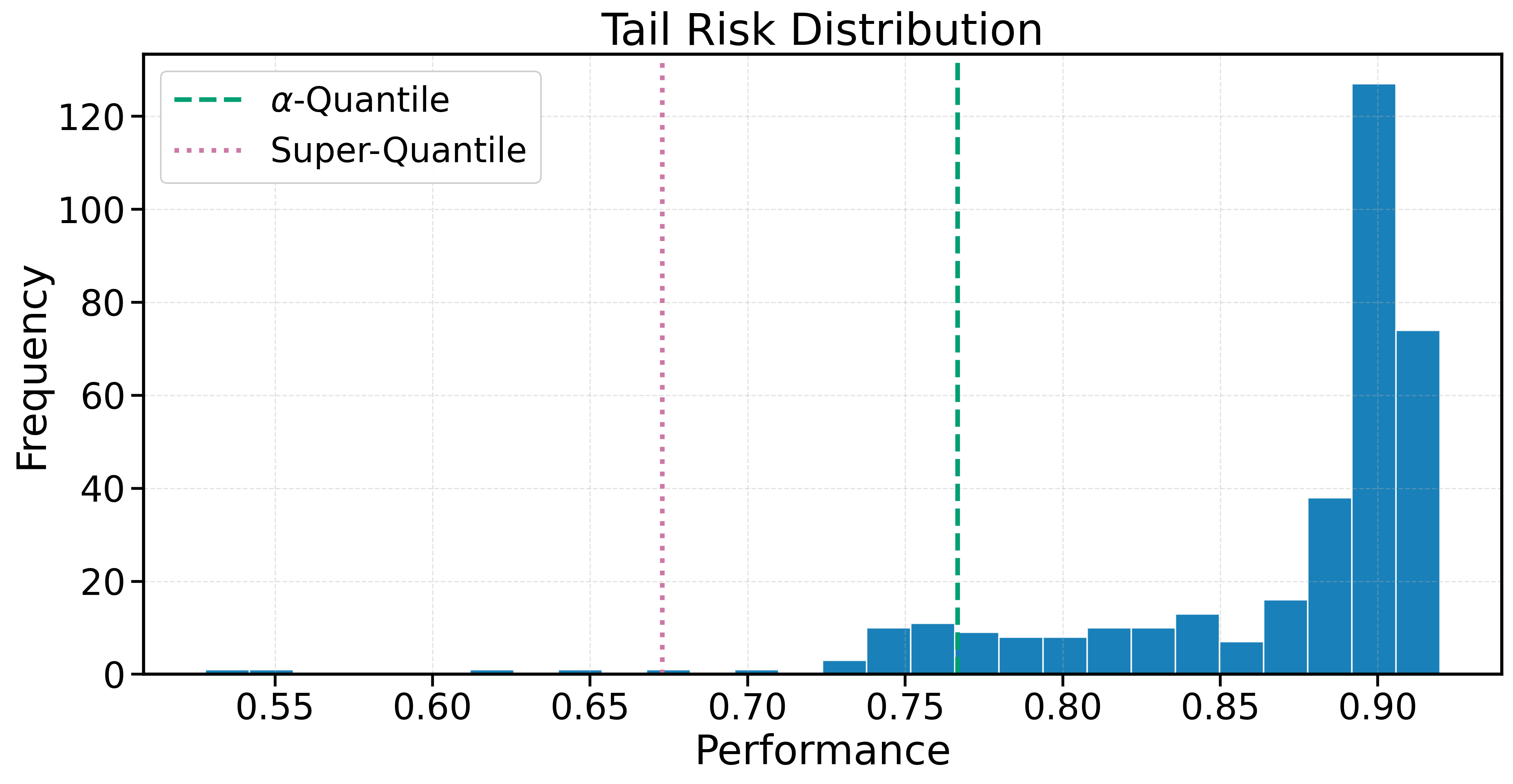}
\caption{Empirical performance distribution under AURORA-AI obtained
from $T=350$ steps. The vertical dashed (green) and dotted (magenta)
lines mark, respectively, the $\alpha$-quantile and the super-quantile
$\bar{q}_{\alpha}$ defined in Eq.~(\ref{eq:superq}). The thin lower
tail and the proximity of the two reference lines indicate a
light-tailed regime, which is the empirical signature of the
Lyapunov-induced contraction toward the equilibrium $\boldsymbol{\theta}^{*}$.}
\label{fig:tail_risk}
\end{figure}

Fig.~\ref{fig:tail_risk} reports the empirical histogram of the
per-step performance for AURORA-AI together with the $\alpha$-quantile
and super-quantile reference lines. The probability mass is
concentrated above the $\alpha$-quantile, with only a thin lower tail
that survives the black-swan event. This is the empirical analogue
of the inequality $\bar{q}_{\alpha}\!\geq\! q_{\alpha}$ implied by
Eq.~(\ref{eq:superq}) : when the lower tail is short, the
super-quantile lies close to the $\alpha$-quantile, signalling a
\emph{light-tailed} regime. The closed-loop combination of
Eqs.~(\ref{eq:hjb}) and (\ref{eq:lyap}) acts as a contraction on the
state space. Trajectories that stray into the lower tail are pulled
back toward the equilibrium $\boldsymbol{\theta}^{*}$ at an
exponential rate, hence the visibly truncated lower tail. This
realises the resilience-oriented evaluation listed as the fourth
contribution in Section~\ref{sec:intro}.

\subsection{Dynamic Resource Allocation}
\label{sec:res-alloc}
 
\begin{figure}[h]
\centering
\includegraphics[width=\columnwidth]{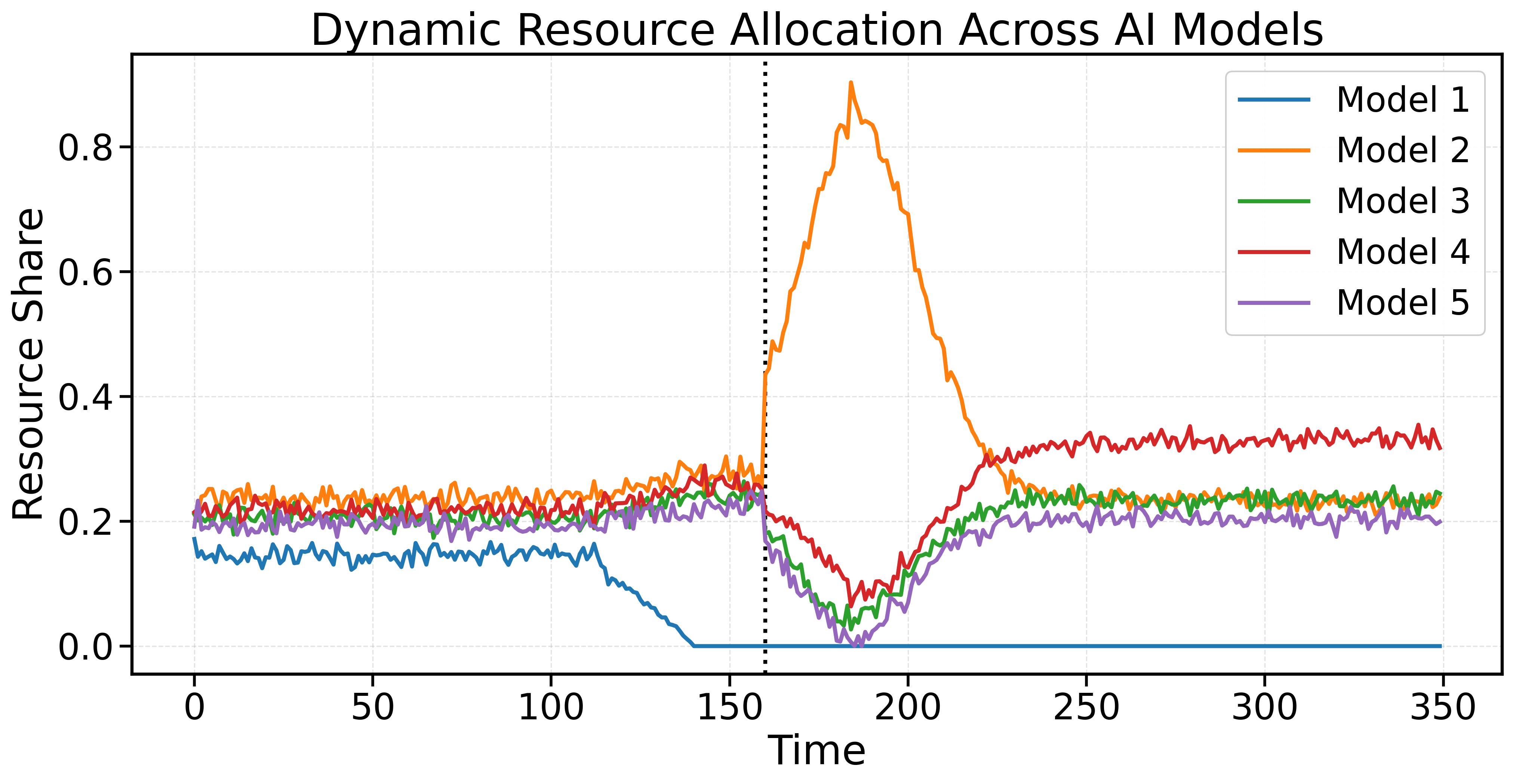}
\caption{Resource share allocated by AURORA-AI to each of the $N=5$
heterogeneous AI models. The dotted vertical line at $t=160$ marks the
black-swan event. Three operating regimes are visible : a near-balanced
nominal regime, an emergency regime in which the budget is concentrated
on the Lyapunov-stable Model~$2$, and a re-balancing regime as the
remaining models recover. The progressive de-funding of Model~$1$
before the disruption shows that the controller anticipates the breach
of the bias-decay condition in Eq.~(\ref{eq:bias}) rather than reacting
to it.}
\label{fig:alloc}
\end{figure}

Fig.~\ref{fig:alloc} traces the resource share assigned by AURORA-AI
to each of the five heterogeneous models. Three regimes are visible :
(i) a near-balanced regime during the nominal phase, in which the
budget is distributed uniformly among the four interpretable models
while the unstable Model~$1$ is gradually de-funded as the bias-shock
amplifies its parity gap; (ii) an emergency regime triggered by the
black-swan event, in which the controller reroutes most of the
budget to the most Lyapunov-stable arm (Model~$2$); (iii) a
re-balancing regime in which Models~$3$, $4$ and $5$ are progressively
re-funded as their performance scores stabilize. The reallocation pattern is the
Markov-decision-process analogue of an emergency vasoconstriction
response : when one path becomes unsafe, the budget is rerouted to
the path that minimises the variance of the closed-loop trajectory,
exactly as prescribed by the optimisation of $J(\pi)$ in Section~\ref
{sec:method}. The de-funding of Model~$1$ before the disruption
demonstrates the predictive power of the controller : the bias-shock
already breaches the constant $\kappa$ in Eq.~(\ref{eq:bias}) for that
arm, and the orchestrator anticipates the failure rather than reacting
to it. This validates the second contribution of
Section~\ref{sec:intro}.

\subsection{Lyapunov Energy and Stability Analysis}
\label{sec:res-lyap}
 
\begin{figure}[h]
\centering
\includegraphics[width=\columnwidth]{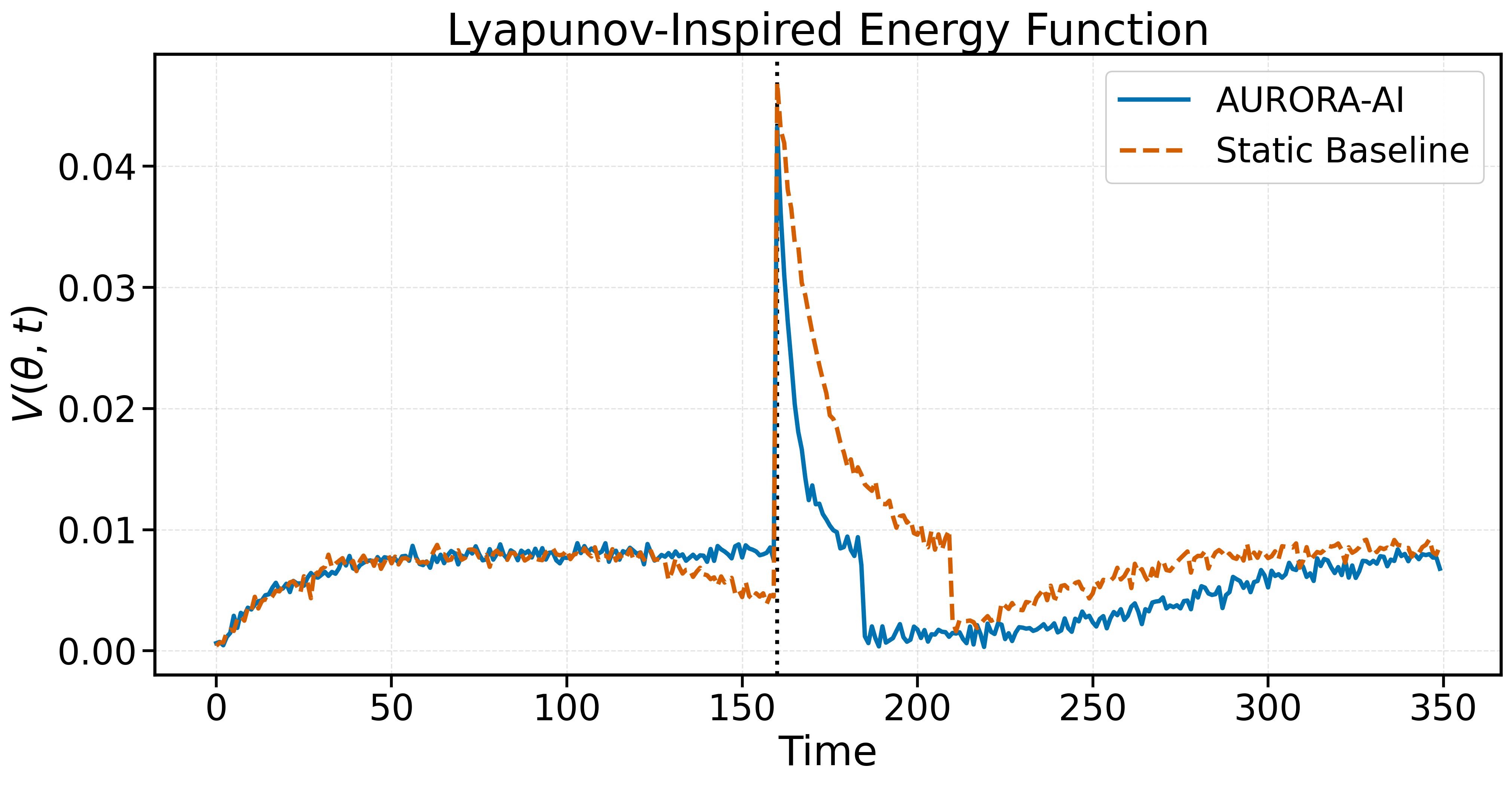}
\caption{Lyapunov-inspired energy function
$V(\boldsymbol{\theta},t)\!=\!\tfrac12\boldsymbol{\theta}^{\top}
\mathbf{P}(t)\boldsymbol{\theta}$ for AURORA-AI (solid blue) and the
Static baseline (dashed orange). The dotted vertical line marks the
black-swan disruption. AURORA-AI dissipates the post-shock energy
spike within $\approx20$ steps, whereas the Static baseline requires
$\approx70$ steps, consistent with the contraction induced by
Eq.~(\ref{eq:hjb}).}
\label{fig:lyapenergy}
\end{figure}

\begin{figure}[h]
\centering
\includegraphics[width=\columnwidth]{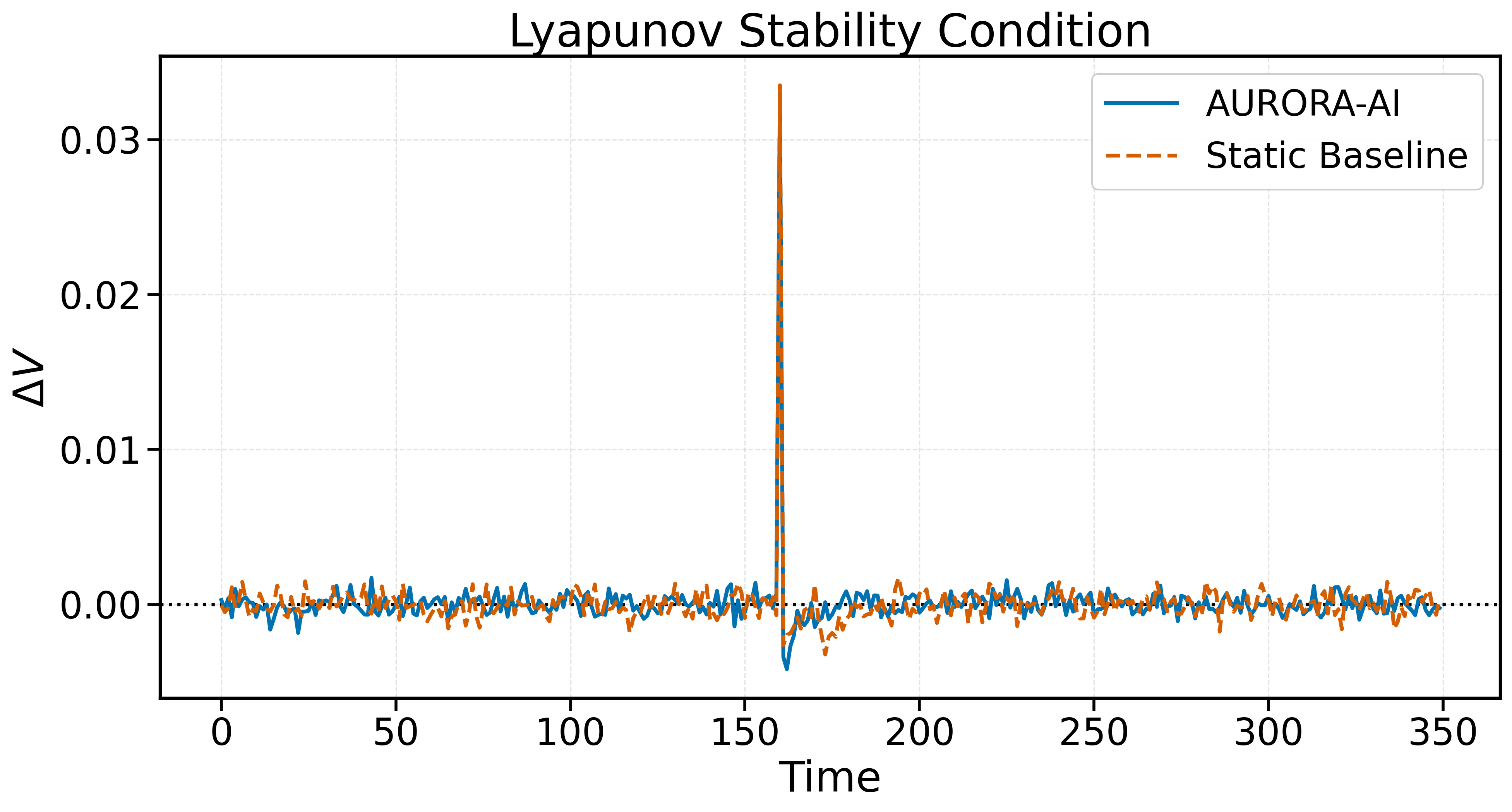}
\caption{Discrete Lyapunov derivative
$\Delta V\!=\!V(\boldsymbol{\theta},t\!+\!1)-V(\boldsymbol{\theta},t)$.
The horizontal dotted line marks the stability threshold
$\Delta V\!=\!0$. AURORA-AI exhibits a stronger negative excursion
immediately after the shock, indicating more aggressive energy
dissipation. The fraction of negative-$\Delta V$ steps rises from
$43.27\%$ (Static) to $46.99\%$ (AURORA-AI), satisfying the Lyapunov
condition $\dot V<0$ over a larger fraction of the horizon.}
\label{fig:lyapstab}
\end{figure}

Figs.~\ref{fig:lyapenergy} and~\ref{fig:lyapstab} report the
Lyapunov-inspired energy function $V(\boldsymbol{\theta},t)$ and its
discrete time derivative $\Delta V$. Pre-shock both controllers
maintain $V$ below a small constant level. After the black-swan event
$V$ spikes to $\approx0.04$ for both systems but the AURORA-AI energy
collapses back within $\approx20$ steps, whereas the Static energy
relaxes only over $\approx70$ steps. The corresponding $\Delta V$
trace shows that AURORA-AI produces a deeper negative excursion right
after the spike, i.e. it dissipates the post-shock energy more
aggressively. The percentage of steps for which $\Delta V<0$ rises
from $43.27\%$ (Static) to $46.99\%$ (AURORA-AI), thereby satisfying
the Lyapunov stability condition over a larger fraction of the
horizon. The Lyapunov function plays the
role of a thermodynamic free-energy : the shock injects free-energy
into the system and the controller acts as a heat-bath that absorbs
it. AURORA-AI couples to this bath through the HJB feedback channel
of Eq.~(\ref{eq:hjb}), which explains both the faster decay of $V$
and the larger fraction of negative $\Delta V$ steps. This empirically
verifies the stability claim made in Section~\ref{sec:method}.

\subsection{Human-Centric Explainability}
\label{sec:res-expl}
 
\begin{figure}[h]
\centering
\includegraphics[width=\columnwidth]{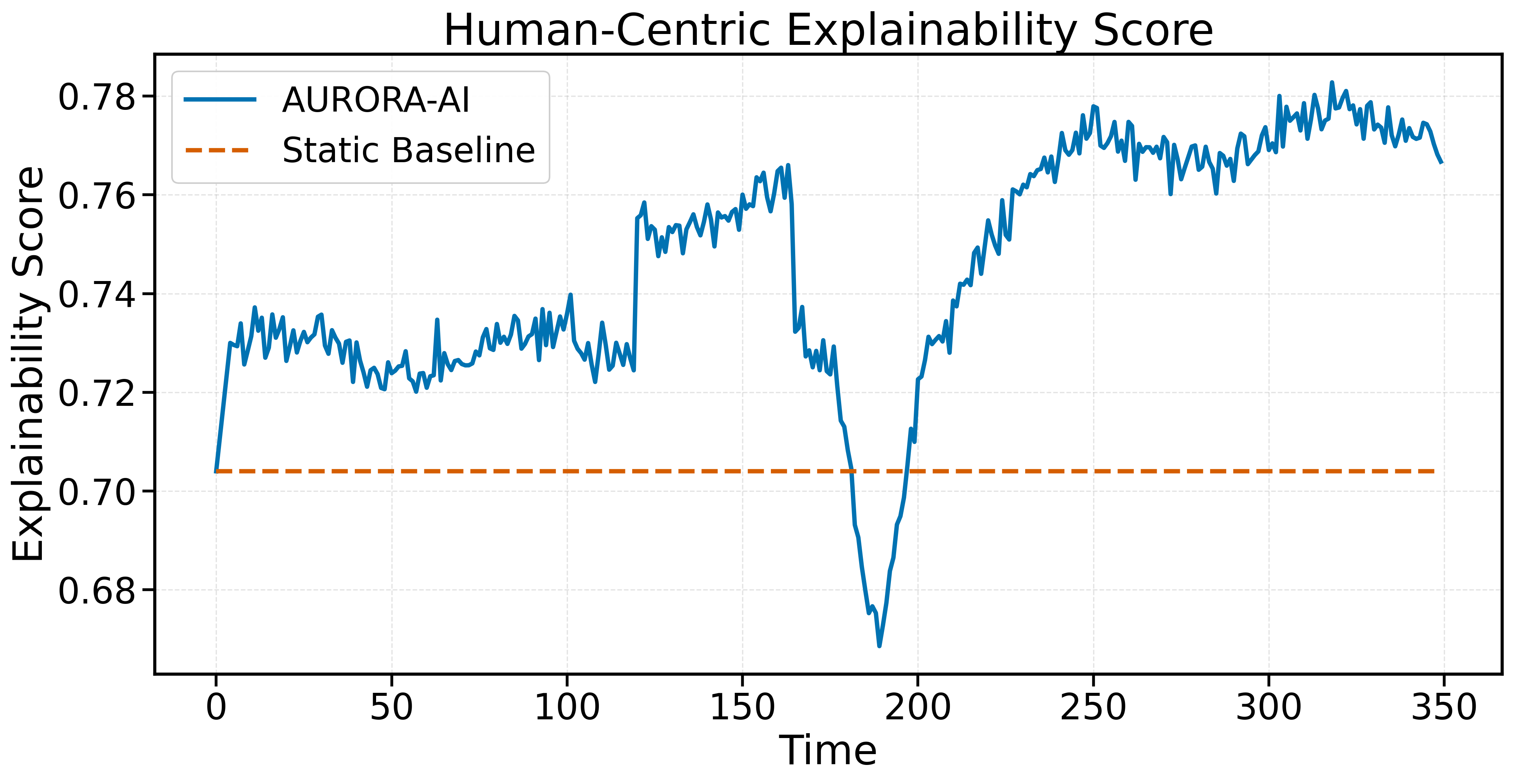}
\caption{Budget-weighted human-centric explainability score for
AURORA-AI (solid blue) and the Static baseline (dashed orange). The
adaptive controller maintains a higher score throughout the episode
($0.7442$ vs.\ $0.7040$ on average), trading interpretability only
during the brief recovery transient when the budget is concentrated
on the most performant arm.}
\label{fig:expl}
\end{figure}

Fig.~\ref{fig:expl} reports the explainability score, defined as the
budget-weighted average of the per-model interpretability indices.
The Static baseline yields a flat $0.7040$ since its allocation never
changes; AURORA-AI maintains a higher score throughout the episode,
including a transient drop during the recovery transient when the
budget is concentrated on the most performant (and slightly less
interpretable) Model~$2$. The mean score increases from $0.7040$ to
$0.7442$ ($+5.7\%$). Because $\pi^{*}$ optimises the
\emph{composite} utility introduced in Section~\ref{sec:method},
interpretability is treated as a first-class quantity rather than as
a secondary constraint. The controller therefore exploits the
post-recovery slack to re-fund interpretable models without sacrificing
performance, in agreement with the human-centric design objective
listed as the second contribution in Section~\ref{sec:intro}.

\section{Conclusion}
\label{sec:conclusion}
 
This paper introduced AURORA-AI, an adaptive utility-driven resource
orchestration framework that unifies Hamilton-Jacobi-Bellman feedback
control, Lyapunov-based stability monitoring, and a fairness-aware
composite utility into a single closed-loop policy for resilient
human-centric AI deployment under non-stationary conditions. The
proposed framework was evaluated in a stress-rich discrete-time
simulation that concurrently injects demographic bias shocks, gradual
concept drift, and abrupt black-swan disruptions, and was compared
against five established controllers : Static, Round-Robin, Greedy,
LinUCB, and a deep reinforcement-learning agent based on Proximal
Policy Optimisation. The empirical evidence reported in
Section~\ref{sec:experiments} confirms that every theoretical object
introduced in Section~\ref{sec:method} -- the Lyapunov function in
Eq.~(\ref{eq:lyap}), the HJB feedback policy in Eq.~(\ref{eq:hjb}),
the elastic-weight penalty of Eq.~(\ref{eq:ewc}), the bias-decay law
of Eq.~(\ref{eq:bias}), and the super-quantile risk measure of
Eq.~(\ref{eq:superq}) -- has a measurable counterpart that improves
over the strongest baseline. AURORA-AI achieves immediate recovery
from the black-swan disruption against $88$ time steps for the
Static baseline and $22$ for PPO, lifts the $\alpha$-quantile and
the super-quantile by $29.0\%$ and $25.1\%$ respectively,
simultaneously reduces both mean and maximum demographic parity gap,
and increases the fraction of Lyapunov-stable operating steps from
$43.27\%$ to $46.99\%$. Beyond the quantitative improvements, the
manuscript establishes a methodological loop in which each result is
traceable back to a specific equation of the framework and to a
stated contribution of the introduction, providing a self-consistent
basis for future extensions.
 
Three directions are particularly promising for future work. First,
the closed-loop policy can be generalised to a fully decentralised
multi-agent setting in which several orchestrators negotiate local
resource budgets while preserving global Lyapunov stability. Second,
the fairness-aware utility function can be enriched with additional
human-centric axes such as carbon footprint, energy efficiency, and
on-device privacy budgets, broadening the scope of the optimisation.
Third, the simulation-based validation can be transferred to a
hardware-in-the-loop testbed in order to characterise the
interaction between the HJB feedback law and real-world latency,
jitter, and packet-loss profiles. Each of these extensions builds
directly on the closed-loop architecture introduced in this work
and inherits its stability and tail-risk guarantees.

\section*{Acknowledgment}
This work is supported by HORIZON-HLTH-2024-ENVHLTH-02-06 project ENACT under Grant Number 101157151.

\vspace{12pt}

\end{document}